\pdfoutput=1

\documentclass[11pt]{article}

\usepackage{acl}

\usepackage{times}
\usepackage{latexsym}

\usepackage[T1]{fontenc}

\usepackage[utf8]{inputenc}

\usepackage{microtype}

\usepackage{silence}
\usepackage{inconsolata}
\usepackage{url}
\usepackage{amsfonts}
\usepackage{booktabs}
\usepackage{multirow}
\usepackage{amsmath}
\usepackage{hhline}
\usepackage{colortbl}
\usepackage{enumerate}
\usepackage[normalem]{ulem}
\usepackage{lipsum}
\usepackage{pgf-pie}
\usepackage{tikz,pgfplots,pgfplotstable}
\pgfplotsset{compat=newest}
\usepackage{verbatim}
\usepackage{ctable}

\usepackage[linecolor=orange,size=scriptsize]{todonotes}
\usepackage{caption}
\usepackage{graphicx}
\usepackage{siunitx}
\usepackage{tabularx}

\usepackage{enumitem}
\setitemize{noitemsep,topsep=2pt,parsep=2pt,partopsep=2pt,leftmargin=*}

\usepackage{amsmath}
\usepackage{amsfonts}
\usepackage{amssymb}
\usepackage{wrapfig}
\usepackage{subcaption}
\usepackage{multirow}
 \usepackage{mathtools} 

\usepackage{verbatim}

\usepackage{anyfontsize}

\usepackage{color}
\usepackage{tikz}
\usetikzlibrary{arrows,shapes,snakes,automata,backgrounds,fit,petri}
\usepackage{adjustbox}

\newcommand{\RN}[1]{%
	\textup{\lowercase\expandafter{\it \romannumeral#1}}%
}

\usepackage{csquotes}
\usepackage{booktabs}

\usepackage{tcolorbox}
\usepackage{colortbl}

\usepackage{xcolor}
\definecolor{mygreen}{HTML}{3cb44b}
\colorlet{myyellow}{green!10!orange!90!}
\makeatletter
\newcommand{\distas}[1]{\mathbin{\overset{#1}{\kern\z@\sim}}}%

\usepackage{enumitem}

\usepackage[lined,boxed,commentsnumbered,ruled,linesnumbered]{algorithm2e}

\newcommand{\ie}[0]{\emph{i.e., }}

\newcommand{\eg}[0]{\emph{e.g., }}

\newcommand{\beq}{\vspace{0mm}\begin{equation}}
\newcommand{\eeq}{\vspace{0mm}\end{equation}}
\newcommand{\beqs}{\vspace{0mm}\begin{eqnarray}}
\newcommand{\eeqs}{\vspace{0mm}\end{eqnarray}}
\newcommand{\barr}{\begin{array}}
\newcommand{\earr}{\end{array}}

\ifx\assumption\undefined
\newtheorem{assumption}{Assumption}
\fi

\ifx\definition\undefined
\newtheorem{definition}{Definition}
\fi

\ifx\remark\undefined
\newtheorem{remark}{Remark}
\fi

\WarningsOn

\newcommand{\model}{\textsc{Godel}}
\newcommand{\fone}{$\mathtt{F_1^R}$} 
\newcommand{\kfone}{$\mathtt{F_1^K}$} 
\newcommand{\smarker}[1]{\tmark[{\makebox[0pt][l]{#1}}]}

\newcommand{\modelp}[2]{\textsc{#1}{$_{\mathtt{#2}}$}}
\newcommand{\modelbase}{\textsc{\model}$_{\mathtt{B}}$}
\newcommand{\modellarge}{\textsc{\model}$_{\mathtt{L}}$}
\newcommand{\modelxlarge}{\textsc{\model}$_{\mathtt{XL}}$}

\newcommand{\modelvone}{\textsc{DialoGPT}}

\newcommand{\smodel}{\textsc{Godel}}
\newcommand{\smodelbase}{\textsc{\smodel}$_{\mathtt{B}}$}
\newcommand{\smodellarge}{\textsc{\smodel}$_{\mathtt{L}}$}
\newcommand{\smodelxlarge}{\textsc{\smodel}$_{\mathtt{XL}}$}

\newcommand{\smodelxlargegptthree}{\textsc{\smodel}$_{\mathtt{XL}}^{\mathtt{GPT-3}}$}
\newcommand{\smodelxlargegptj}{\textsc{\smodel}$_{\mathtt{XL}}^{\mathtt{GPT-J}}$}

\newcommand{\smodelvonebase}{\textsc{DGPT}$_{\mathtt{B}}$}
\newcommand{\smodelvonelarge}{\textsc{DGPT}$_{\mathtt{L}}$}

\definecolor{darkred}{rgb}{0.55, 0.0, 0.0}

\title{\model{}: Large-Scale Pre-Training for Goal-Directed Dialog}

\author{
Baolin Peng\textsuperscript{$\dagger$} \hspace{.1cm} 
\textbf{Michel Galley}\textsuperscript{$\dagger$} \hspace{.1cm}
\textbf{Pengcheng He}\textsuperscript{$\dagger$} \hspace{.1cm} 
\textbf{Chris Brockett}\textsuperscript{$\dagger$}\\
\textbf{Lars Liden}\textsuperscript{$\dagger$} \hspace{.1cm}
\textbf{Elnaz Nouri}\textsuperscript{$\dagger$} \hspace{.1cm}
\textbf{Zhou Yu}\textsuperscript{$\ddagger$} \hspace{.1cm}
\textbf{Bill Dolan}\textsuperscript{$\dagger$} \hspace{.1cm}
\textbf{Jianfeng Gao}\textsuperscript{$\dagger$} \\
\textsuperscript{$\dagger$} Microsoft Corp. \textsuperscript{$\ddagger$} Columbia University\\
\texttt{\{bapeng,mgalley,penhe,chrisbkt,laliden,elnouri,billdol,jfgao\}@microsoft.com} \hspace{.1cm}\\
\texttt{zy2461@columbia.edu}
}

\begin{document}

\maketitle
\begin{abstract}
We introduce \model{} ({\bf G}rounded {\bf O}pen {\bf D}ialogu{\bf e} {\bf L}anguage Model), a large pre-trained language model for dialog. In contrast with earlier models such as DialoGPT, \model{} leverages a new phase of {\it grounded} pre-training designed to better support adapting \model{} to a wide range of downstream dialog tasks 
that require information external to the current conversation (\eg a database or document) to produce good responses. Experiments against an array of benchmarks that encompass task-oriented dialog, conversational QA, and grounded open-domain dialog show that \model{} outperforms state-of-the-art pre-trained dialog models in few-shot fine-tuning setups, in terms of both human and automatic evaluation. A novel feature of our evaluation methodology is the introduction of a notion of utility that assesses the {\it usefulness} of responses (extrinsic evaluation) in addition to their communicative features (intrinsic evaluation). We show that extrinsic evaluation offers improved inter-annotator agreement and correlation with automated metrics. Code and data processing scripts are publicly available.\footnote{\url{http://aka.ms/GODEL}}
\end{abstract}

\section{Introduction}
\label{section:Introduction}
This work describes the development of a very large pre-trained dialog model -- {\bf G}rounded {\bf O}pen {\bf D}ialogu{\bf e} {\bf L}anguage Model (\model{}). As the name indicates, \model{} is designed for general-domain conversation and is fully open-sourced. \model{} should be of technical interest for two reasons. First, it is pre-trained in three phases, successively folding in data from web text, publicly-available dialog (\eg Reddit), and a collection of existing corpora that support grounded dialog tasks. The grounded dialog corpora, which include MS MARCO \cite{nguyen2016ms} and DSTC7 \cite{DSTC7}, allow for more effective fine-tuning on dialog tasks where responses must be conditioned on information external to the current conversation (\eg a retrieved document.) Second, \model{} is validated on a utility-driven suite of benchmarks specifically designed for few-shot fine-tuning of {\it open-ended goal-directed general-domain dialog} models. We will show that \model, as validated using this methodology, is more readily amenable to fine-tuning for goal-directed dialog tasks than other large pre-trained language models.

Our approach seeks to address a long-standing obstacle to general-purpose open-ended conversation models, namely a lack of robust automated evaluation criteria that can drive development~\citep{GaoEtAl:2019}. Recent state-of-the-art models that leverage large PLMs \cite[\eg][]{zhang2019dialogpt,DeFreitas2020TowardsAH,Roller2021RecipesFB,Bao2021PLATOXLET,Thoppilan2022LaMDALM,gao2022neural} offer the potential for substantive open-ended conversational interactions, yet they resist meaningful comparison owing to the lack of consensus on evaluation. 

This poses a fundamental question: what do we want of a good general-purpose dialog model in the first place? We take it as a given that it should be fluent and socially engaging. Indeed, most SOTA PLMs are primarily evaluated on such \textit{intrinsic} communicative dimensions. But beyond that, we must also acknowledge that machine-human conversation typically serves a purpose and aims to fulfill one or more goals on the part of the user. In other words, the model must offer \textit{utility} to the user. It is this \textit{extrinsic} dimension of functional utility, we suggest, that constitutes the proper focus of automated evaluation in general-domain models.

In the second half of this paper (Sections~\ref{section:Results} and~\ref{section:Analyzes}), we explore this notion of Utility in fine-tuning \model{} on four established tasks that cover multiple domains and conversational settings, ranging from the task-oriented MultiWOZ \cite{budzianowski2018multiwoz} to more open-ended goal-oriented tasks, \ie CoQA \cite{reddy2019coqa}, Wizard of Wikipedia \cite{dinan2018wizard}, and Wizard of the Internet \cite{komeili2021internet}. We show that \model{} improves on a baseline \modelvone{} \cite{zhang2019dialogpt}, with performance gains in goal-directed tasks. Our results on these diverse tasks speak to the effectiveness of our approach. We also find that the focal shift to goal-directedness and extrinsic evaluation affords better inter-rater agreement and human-metric correlations.

As baselines for future research, we release three versions of \model{}: small (\modelbase{}) and large (\modellarge{}) initiated from T5 and T5-Large respectively, and a very large model (\smodelxlargegptj{}) that is trained off GPT-J~\cite{gao2020pile,gpt-j}.\footnote{The discussion of our largest model in this paper relates to \modelxlarge{} which is based on GPT-3, but which we are unable to release owing to licensing restrictions. We therefore release \smodelxlargegptj{} a substitute model based on GPT-J that performs similarly. A comparison of GODEL models initialized from GPT-3 and GPT-J is provided in the Appendix.} We also furnish code to download datasets for fine-tuning and evaluation. The release is fully open sourced (models, data, and code) and tightly integrated with the latest version of HuggingFace's transformers library.

\section{Related Work}
\label{section:RelatedWork}
Large pre-trained language models (PLMs) for dialog \cite{zhang2019dialogpt,DeFreitas2020TowardsAH,Roller2021RecipesFB,Peng2021SoloistBT,Bao2021PLATOXLET,Thoppilan2022LaMDALM} have had significant impact in Conversational AI, as these models have brought strong performance gains in competitions such as DSTC \cite{Gunasekara2020OverviewOT} and ConvAI \cite{ConvAI3}, and have been widely adopted in both task-oriented and chitchat research. While the applications of these dialog models are numerous, empirical evaluation of dialog pre-training has mostly been restricted to {\it intrinsic} evaluation (\ie agnostic to any notion of utility). For example, DialoGPT \cite{zhang2019dialogpt} offered Reddit-based pre-trained models that are evaluated on Relevance, Informativeness, and Humanness and various string overlap based automated evaluation metrics. Meena \cite{DeFreitas2020TowardsAH}'s evaluation is focused on Sensibleness and Specificity, while also offering a new automated evaluation (SSA) that measures these two qualities. BlenderBot \cite{Roller2021RecipesFB} contributes a variety of models that capitalize on skills including personality, empathy, and knowledge, but evaluation in this work centers mostly on measures of Humanness and Engagingness. 

Plato-XL \cite{Bao2021PLATOXLET} focuses on intrinsic evaluation with measures of Coherence, Inconsistency, Informativeness, Hallucination, and Engagingness. Plato-XP does however provide some extrinsic evaluation on DSTC9-Track1 \cite{kim2020beyond}, MultiWOZ 2.2 \cite{zang-etal-2020-multiwoz}, and DuConv \cite{Wu2019ProactiveHC}, but is based on task-specific automated metrics (ROUGE-L for DSTC9, Goal-Accuracy for MultiWOZ, and Knowledge-F1 for DuConv). By contrast, our work provides a more unified measure of Utility, which is measured the same fashion (\eg same annotation instructions) across dialog tasks and datasets. Finally, LaMDA \cite{Thoppilan2022LaMDALM} presents dialog models motivated by dialog applications, but is evaluated primarily using intrinsic measures (Sensibleness, Specificity, Interestingness, Safety, Groundedness). In its human evaluation, LaMDA does consider an extrinsic evaluation measure of Utility (Helpfulness), and is probably the closest to our work in terms of analyzing PLMs with a unified extrinsic evaluation. Our works seeks to go further by contrasting intrinsic and extrinsic measures, and analyzing their correlates with automatic evaluation on specific tasks and datasets.

\section{Open-Domain Goal-Directed Dialog}
\label{section:Framework}
In the goal-directed framework of this paper, we seek to generate human-like conversational responses that attain a high level of {\it Utility}. While prior work in task-oriented dialog has used various measures of Utility (\eg Inform-rate and Success-rate for MultiWOZ~\cite{budzianowski2018multiwoz}, Knowledge-F1 for Wizard of Wikipedia \cite{dinan2018wizard}), these ad-hoc measures are often heterogeneous and only applicable to a limited number of subtasks. This makes it difficult to analyze results across dialog tasks and datasets. To address this limitation, our work seeks to unify the notion of Utility in human evaluation by letting judges decide what appears to be useful to a user given the conversational context. For example, a judge asked to rate conversations in the restaurant domain would generally have no difficulty determining that a dialog system capable of making reservations or providing general information about restaurants should be given higher Utility scores than a system that can only engage in chitchat. The generality of Utility is comparable to that of other qualities such as Humanness and Interestingness \cite{Li2019ACUTEEVALID} often used in chitchat work, but unified evaluation of this kind is rarely used in more task-oriented dialog research. Section~\ref{section:Analyzes} will analyze and contrast both intrinsic (\eg Humanness) and extrinsic (\eg Utility) qualities measured under human evaluation, and correlate them with automatic evaluation metrics.

Considering the goal of generating responses that attain high Utility, we cast the {\it open-domain goal-directed dialog} generation task as follows: Given a source dialog context $S = (s_1, \cdots , s_N)$ and an environment $E$, the goal is to generate a target sentence $T=(t_{1}, \cdots , t_N)$. The conditional probability of $P(T|S, E)$ can be written as the product of a series of conditional probabilities:
\begin{equation}
p(T|S, E) = \prod_{n=1}^N p(t_n|t_1, \cdots, t_{n-1}, S, E)
\label{eq:lm}
\end{equation}
The environment $E$ represents the state of the world and external knowledge (\eg a database or results of a search engine) that are needed to go beyond chitchat and satisfy a need for Utility. In the pre-training stage, $E$ is generally absent as $E$ tends to be task specific, but we find it effective to substitute $E$ with a grounding text during the third phase of pre-training, \ie grounded dialog pre-training.

\section{Models}
\label{section:Models}
\begin{figure}[t]
\centering
\includegraphics[width=1.0\columnwidth]{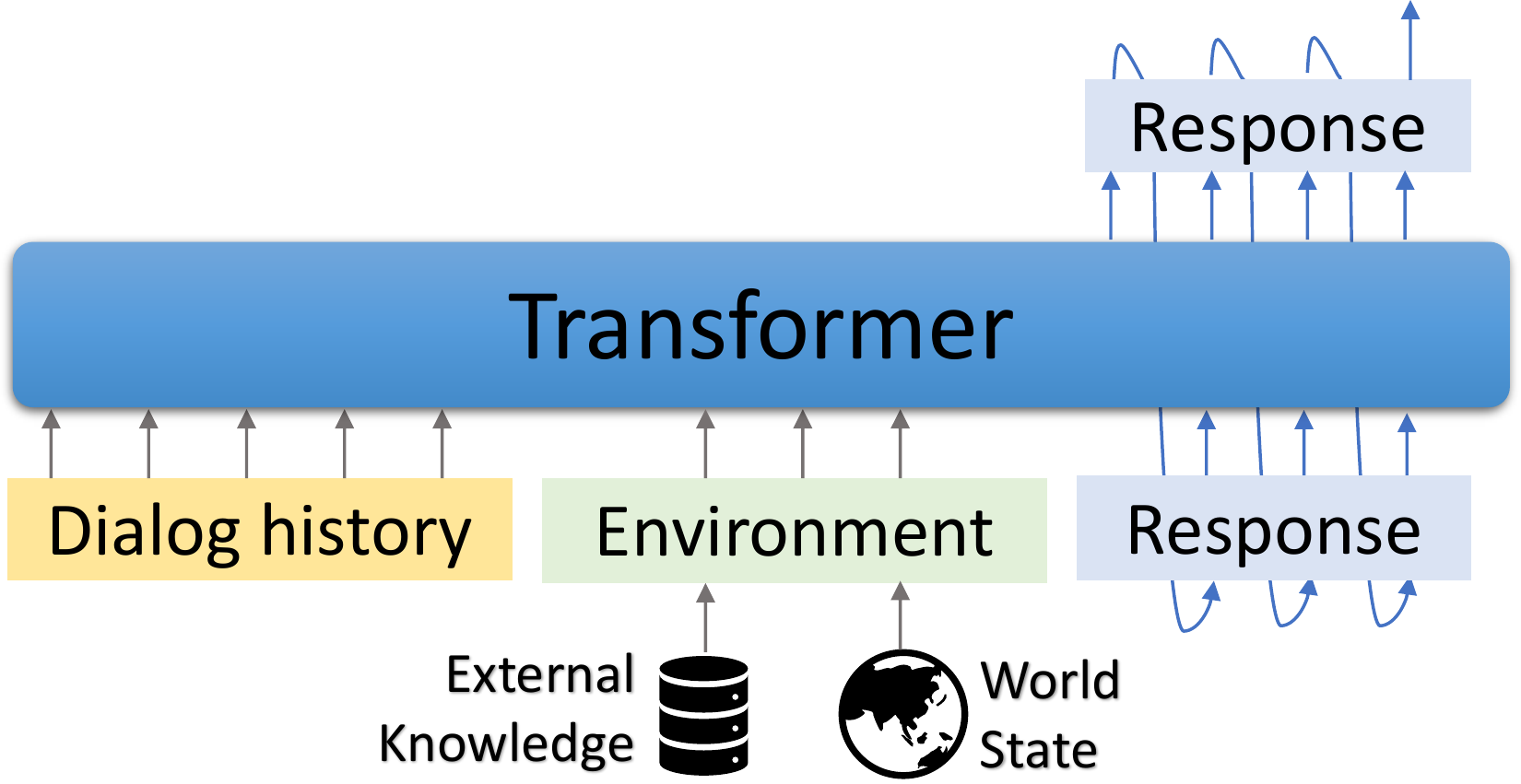}
\caption{\model{} pre-training and fine-tuning with a Transformer-based encoder-decoder model, taking dialog context and environment (world state or external knowledge) as an input represented as a string.}
\label{fig:model_architecture}
\end{figure} 

At its core, \model{} is a standard pre-trained LM, trained on web text data. We utilize a sequence-to-sequence Transformer model, as shown in Figure \ref{fig:model_architecture}, to generate responses given dialog history and environment. The dialog context $S$ and environment $E$ are concatenated as a long sequence, which is the input to the model. A sample training example is shown in Figure~\ref{fig:example}.

\begin{figure}[t]
\begin{tcolorbox}[boxsep=1pt,colframe=white!90!black,colback=black!2!white] \small
 \textcolor{red!60}{User : I would like to find an expensive restaurant that serves Chinese food. System : sure, which area do you prefer ? User : Bellevue downtown.} 
 \textcolor{blue}{<|environment|> Multiple expensive Chinese restaurants in Bellevue downtown. =>}
 \textcolor{mygreen}{There are multiple restaurants meet your requirement. peony kitchen is a great Chinese Restaurant. Would you like to book a table there?}
\end{tcolorbox} 
\caption{Sample training instance, with conversation history in red, grounding in blue, and response in green.}
\label{fig:example}
\end{figure}

\model{} is pre-trained in three phases: (1) Linguistic pre-training on public web documents to provide a basic capability for text generation. (2) Dialog pre-training on public dialog data to improve the models' handling of general conversational behavior. (3) Grounded dialog pre-training to enable grounded response generation.
 
For general-domain dialog pre-training, we use a dataset extracted from Reddit comment chains for DialoGPT \cite{zhang2019dialogpt}, which consists of 147M dialog sessions for a total of 6B tokens. For the final grounded dialog pre-training phase, we use a corpus constructed by leveraging existing datasets that support knowledge grounded response generation, conversational question-answering, and task-oriented dialog, as follows:
 
 \begin{itemize}
 \item The DSTC7 Task 2 corpus~\cite{galley2018end} is designed for end-to-end conversational modeling that goes beyond chitchat by injecting informational responses grounded in external knowledge. This has 2 million examples for training.
 \item MS MARCO~\cite{nguyen2016ms} is a large-scale question-answering dataset. Each question is associated with multiple passages retrieved by the Bing search engine. Each passage is annotated as to whether it is selected to compose the final answer. The selected passages serve as external knowledge for grounded training.
 \item UnifiedQA~\cite{khashabi2020unifiedqa} is a large-scale question-answering corpus that unifies different QA datasets into a plain text-to-text format. Similarly, context paragraphs are treated as the external knowledge on which the model is trained to generate answers given questions.
 \item The Schema-Guided Dialog~\cite{chen2020schema} is a multi-domain, task-oriented conversation dataset that contains approximately 20k annotated examples. Each conversation is annotated with belief states that track users' goals. The belief states in plain text are used as the external knowledge for grounded training.

\end{itemize}

We consider three sizes of model: a 220M parameter base version (\modelbase{}), a 770M parameter version (\modellarge{}), and a 175B parameter version (\modelxlarge{}). The 220M parameter base model (\modelbase{}) has a 12 layer encoder and 12 layer decoder with 768 dimensional embeddings. The 770M parameter model (\modellarge{}) doubles the number of layers for encoder and decoder and has 1024 dimensional embeddings. These two models, \modelbase{} and \modellarge{}, are initiated from T5 and T5-Large, respectively, and are based on the HuggingFace repository versions. We use Byte-Level BPE tokenization as implemented in HuggingFace. 

\modelxlarge{} is initiated from GPT-3 and is based on the implementation by~\citet{NEURIPS2020_1457c0d6}. We chose GPT-3 for our initialization of \modelxlarge{}, despite being aware that we would not be able to release a model due to licensing restrictions, because we wanted to test our grounded fine-tuning recipe on one of the best pre-trained language models available at the time of the writing. As a proxy for the GPT-3-initialized \modelxlarge{}, we release a version pre-trained from GPT-J. Table~\ref{table:results_gptj} in the Appendix shows that this substitute model is competitive with the GPT-3-based version.
 
\modelbase{} and \modellarge{} were trained on 16 Nvidia V100 machines, and \modelxlarge{} was trained with 128 Nvidia V100 GPUs. Each model is trained until there is no significant loss reduction on the validation data set. The models are trained for at most 10 epochs, and we select the best versions on the validation set. We use beam search for inference using a beam size of 5.

\section{Experiments}
\label{section:Results}
\subsection{Experimental Setup}

We evaluate fine-tuned models in two different setups: few-shot and full. Our evaluation focuses primarily on few-shot finetuning, for the following reasons. First, datasets of goal-directed dialogs (\eg MultiWOZ) are more expensive to construct and tend to be much smaller than chitchat datasets, hence we want to assess models’ ability to perform well even if fine-tuning examples are relatively few. Second, this work focuses on language model pre-training for dialog, and evaluation in few-shot setups more directly evaluates the effectiveness of our released {\it pre-trained} models. Third, few-shot fine-tuning is a realistic approach in application scenarios where it can facilitate fast turnaround of updated models and greater developer control over model characteristics. 

\model{} can be used as an initial model to fine-tune for any open-domain goal-directed dialog tasks with a handful of annotated examples. We evaluate \model{} in terms of its success in fine-tuning three types of goal-directed dialog, \ie knowledge-grounded response generation, task-oriented dialog, and conversational QA:

\paragraph{Evaluation datasets:}
\begin{itemize}
	\item Wizard of Wikipedia~\cite{dinan2018wizard} is a dataset of multi-turn knowledge grounded dialogs between an apprentice and a wizard, who has access to Wikipedia sentences and labels the sentences utilized for each utterance they made. To focus on grounded response generation, we use the gold Wikipedia sentences provided at each turn of the dialog.
	\item Wizard of Internet~\cite{komeili2021internet} is an open-domain dialog dataset grounded on internet retrieved sentences. At each turn, the wizard can issue a free text web search and replies to the apprentice grounding the response on the retrieved sentences. Similarly, we use the gold retrieved sentences provided at each turn of the dialog.
	\item MultiWOZ~\cite{budzianowski2018multiwoz} is a multi-domain task-oriented dialog dataset that contains 8438/1000/1000 dialogs for training/validation/testing, respectively. Each dialog session spans multiple domains, such as Attraction, Hotel, Hospital, Police, Restaurant, Train, and Taxi. This dataset is challenging owing to its multi-domain setting and diverse language styles. For this dataset, we consider the task of context-to-response, \ie the model generates responses given the golden belief state and database status in plain text.
	\item CoQA\cite{reddy2019coqa} is a conversational question answering dataset that contains 127,000 questions and answers collected from seven diverse domains. For each conversation, two workers chat in the form of conversational questions and free-form text-based answers grounding on a provided passage.
\end{itemize}

\noindent 
These existing goal-oriented dialog corpora contain thousands of dialog instances. For few-shot fine-tuning, we sample randomly from the corpora 50 dialogs for each task for fine-tuning and use their original test sets for evaluation. We fine-tune \model{} on each task with the same setup as in pre-training, using the training objective shown in Equation~\ref{eq:lm}. The best model is selected based on the perplexity score on the validation set.

\paragraph{Baseline pretrained models} We compare \model{} against established pretrained models: T5 \cite{t5}, BART \cite{bart}, DialoGPT \cite{zhang2019dialogpt}, and BlenderBot \cite{Roller2021RecipesFB}. For T5, we fine-tune from both T5-base (\modelp{T5}{B}) and T5-large (\modelp{T5}{L}). For BART, we fine-tune from both BART (\modelp{BART}{B}) and BART-large (\modelp{BART}{L}). For BlenderBot, we fine-tune from \modelp{Blenderbot}{400M}, which is distilled from a 3B model.\footnote{We used version 1.0 of BlenderBot. BlenderBot 2.0 is not evaluated in this paper as it is specifically designed for internet augmented generation and uses a paradigm that is different from that of task-oriented dialog (\eg MultiWOZ).}

\paragraph{Automatic evaluation metrics} As surrogates for the broad notion of Utility introduced in Section~\ref{section:Framework}, we use the following functions:

\begin{itemize}
\item \fone{} is a metric measuring the average overlap between the prediction and ground truth response or answer.
 
\item \kfone{}, \ie knowledge F1, measures the overlap between the model's response and the knowledge sentences on which the responses are grounded during dataset collection.
 
\item \texttt{Inform} indicates whether the model provides sufficient information to satisfy users' information seeking needs, as defined for MultiWOZ \cite{budzianowski2018multiwoz}.
\item \texttt{Success} measures whether the model provides appropriate information and answers contain all required attributes, as defined for MultiWOZ \cite{budzianowski2018multiwoz}.
\item The \texttt{Combined} score \cite{mehri2019structured} defined as (\texttt{Inform} + \texttt{Success}) * 0.5 + BLEU is used as an overall quality measure.
\end{itemize}

Besides these utility functions (extrinsic), we report intrinsic evaluation results using BLEU, BLEURT~\cite{sellam-etal-2020-bleurt}, BERTScore~\cite{DBLP:journals/corr/abs-1904-09675}, and chrF~\cite{popovic-2015-chrf}. All BLEU scores in this paper are based on corpus-level BLEU-4. In tables \ref{table:results_fewshot_aggregated} through \ref{table:results_fewshow_coqa}, significance is computed using a paired 2-sided t-test against the best competitor. 

\paragraph{Human evaluation setup}

We perform a turn-level human evaluation to investigate whether \model{} generates responses that are 1) useful, 2) human-like and 3) safe. Using Amazon Mechanical Turk, we hired master-level workers with lifetime HIT acceptance \% $>$ 95. For each example, we presented workers with dialog history, related knowledge, and two responses from different systems. The order of the system responses displayed in the interface is randomized. 
We consider three questions: 

\begin{itemize}
	\item Extrinsic evaluation: Which response sounds more useful? (\ie contributes to making the conversation productive, especially towards achieving any stated goals)
	\item Intrinsic evaluation: Which speaker sounds more human? (\eg coherent, fluent, and natural)
	\item Safety evaluation: Which response is socially safer? (\eg friendly, polite, and empathetic, as opposed to harmful, biased, misinformative, or incomprehensible)
\end{itemize}
These were judged on a 5-point Likert scale. A snapshot of the human evaluation interface is provided in Appendix Figure \ref{fig:human_eval}.

\subsection{Automatic Evaluation Results}
\begin{table}

\centering
\scalebox{0.75}{  
\begin{tabular}{l@{\hskip1pt}c@{\hskip5pt}r@{\hskip11pt}r@{\hskip11pt}r@{\hskip5pt}r@{\hskip11pt}r}
\toprule
Model                               & \#params & \textsc{Bleu}  & \fone{} & \textsc{Bert} & \textsc{Bleurt} & chrF \\
\midrule
\modelp{BART}{B}           & 139M & 7.84 &  34.23 & 86.93 & 40.94 & 27.03 \\
\modelp{T5}{B}          & 220M  & 9.13 & 34.75 & 86.32 & 41.30 & 27.42 \\       
\smodelvonebase{}     & 345M  &     0.39 & 5.56 & 56.86 & 16.76 & 6.28 \\
BlenderBot & 400M & 2.50 & 12.98 & 83.77 & 27.31 & 18.70 \\

\modelp{BART}{L}  & 406M &     9.68 & 37.29 &87.33 &44.43 &30.66 \\
\smodelvonelarge{} & 762M &   5.19 & 23.28 & 86.22 & 33.09 & 21.73 \\
\modelp{T5}{L}  & 770M &     12.84 & 44.47 & 90.02 & 47.86 & 34.99 \\
\hline
\smodelbase{} & 220M & 12.77 & 42.80 & 89.59 & 47.79 & 34.42\\  
\smodellarge{}    & 770M &  \bf{13.47}\smarker{**} & \bf{45.82}\smarker{**} & \bf{90.27}\smarker{*} & \bf{49.37}\smarker{**} & \bf{35.87}\smarker{**} \\
\bottomrule
\end{tabular}
}
\caption{Automated evaluation on all tasks in the few-shot fine-tuning setting. `\textsc{Bert}' is BERTScore. All differences between \smodellarge{} and the closest competitor are statistically significant. 
(\mbox{\textsuperscript{*}$p<0.05$, \textsuperscript{**}$p<0.001$}).
}
\label{table:results_fewshot_aggregated}
\end{table}

\begin{table}
\centering
\scalebox{0.75}{  
\begin{tabular}{l@{\hskip4pt}c@{\hskip8pt}r@{\hskip12pt}r@{\hskip12pt}r@{\hskip8pt}r@{\hskip12pt}r}
\toprule
Model                               & \#params & \textsc{Bleu}  & \fone{} & \textsc{Bert} & \textsc{Bleurt} & chrF \\
\midrule
\modelp{BART}{B}     & 139M &   12.75 & 46.61 & 88.64 & 50.78 & 34.43 \\
\modelp{T5}{B}     & 220M &   16.06 & 48.38 & 90.63 & 51.13 &38.62  \\ 
\smodelvonebase{} & 345M &  10.15 & 30.03 & 87.47 & 37.64  &28.98   \\
\hline
\smodelbase{} & 220M & \bf{16.30}\smarker{*}  & \bf{48.87}\smarker{*} & \bf{90.74}\smarker{*} & \bf{51.73}\smarker{*} & \bf{39.14}\smarker{*} \\
\bottomrule
\end{tabular}
}
\caption{Automated evaluation on all tasks in the full fine-tuning setting. `\textsc{Bert}' is BERTScore. All differences between \smodelbase{} and its closest competitor are statistically significant 
(\mbox{\textsuperscript{*}$p<0.05$}).
}
\label{table:results_full_aggregated}
\end{table}

\addtolength{\tabcolsep}{-2pt}

\begin{table}
\centering
\scalebox{0.8}{  
\begin{tabular}{l@{\hskip15pt}r@{\hskip15pt}r@{\hskip15pt}r@{\hskip15pt}r}
\toprule
Model                               & \textsc{Bert} & \textsc{Bleu}  & \fone{} & \kfone{}  \\
\midrule
\modelp{T5}{B}         & 86.16 & 11.38 & 30.53 & 62.34 \\
\smodelbase{} & \bf{86.77}\smarker{**} & \bf{12.49}\smarker{**} & \bf{32.92}\smarker{**} & \bf{62.74} \\
\hline
\modelp{T5}{L}  & 86.26 & 12.21 & 31.04 & 60.52 \\
\smodellarge{}    & \bf{86.79}\smarker{*} & \bf{12.38} & \bf{32.89}\smarker{*} & \bf{60.88} \\
\bottomrule

\end{tabular}
}
\caption{Automated evaluation on grounded generation tasks in the few-shot fine-tuning setting. \textbf{Bold} is best in class. Statistically significant differences are indicated with asterisks (\mbox{\textsuperscript{*}$p<0.05$, \textsuperscript{**}$p<0.001$}).
} 
\label{table:results_fewshow_wow_woi}
\end{table}

\begin{table}
\centering
\scalebox{0.8}{  
\begin{tabular}{lrrrrr}
\toprule
Model                               & \textsc{Bert} & \textsc{Bleu}  & Inform & Success & Combined  \\
\midrule
\modelp{T5}{B}          &  79.21 & 4.31 & 60.60 & 22.50 & 45.86 \\
\smodelbase{} & \bf{87.58}\smarker{**} & \bf{12.81}\smarker{**} & \bf{67.60}\smarker{**} & \bf{46.10}\smarker{**} & \bf{69.72}\smarker{**} \\
\hline
\modelp{T5}{L}  & 88.14 & 12.69 & 71.50 & 56.20 & 76.54 \\
\smodellarge{}    & \bf{88.34}\smarker{*} & \bf{14.07}\smarker{**} & \bf{81.60}\smarker{**} & \bf{62.10}\smarker{**} & \bf{85.90}\smarker{**} \\
\bottomrule
\end{tabular}
}
\caption{Automated evaluation on MultiWOZ dataset in the few-shot fine-tuning setting. \textbf{Bold} is best in class. Statistically significant differences are indicated with asterisks (\mbox{\textsuperscript{*}$p<0.05$, \textsuperscript{**}$p<0.001$}).
}
\label{table:results_fewshot_multiwoz}
\end{table}
\addtolength{\tabcolsep}{2pt}

\begin{table}
\centering
\scalebox{0.8}{ 
\begin{tabular}{l@{\hskip15pt}r@{\hskip15pt}r@{\hskip15pt}r}
\toprule
Model                               & \textsc{Bert} & \textsc{Bleu}  & \fone{}   \\
\midrule
\modelp{T5}{B}         &93.04 & \bf{43.78} & 64.60 \\
\smodelbase{} & \bf{94.03}\smarker{**} & 43.17 & \bf{72.00}\smarker{**} \\
\hline
\modelp{T5}{L}  & 95.19 & \bf{66.94}\smarker{*} & 79.10 \\
\smodellarge{}    & \bf{95.24} & 64.04 & \bf{79.60} \\
\bottomrule
\end{tabular}
}
\caption{Automated evaluation on CoQA in the few-shot fine-tuning setting. \textbf{Bold} is best in class. Statistically significant differences are indicated with asterisks ( \mbox{\textsuperscript{*}$p<0.05$, \textsuperscript{**}$p<0.001$}).
}
\label{table:results_fewshow_coqa}
\end{table}

\begin{table}
\centering
\scalebox{0.8}{
\begin{tabular}{l|cccccccccccc}
\toprule
  & \multicolumn{12}{c}{MultiWOZ}                                                                                      \\
  \midrule
  & \multicolumn{3}{c}{$\mathtt{Inform}$} & \multicolumn{3}{c}{$\mathtt{Success}$} & \multicolumn{3}{c}{\textsc{Bleu}} & \multicolumn{3}{c}{$\mathtt{Combd}$} \\
  \midrule
GPT-3 & \multicolumn{3}{c}{58.90}       & \multicolumn{3}{c}{9.10}        & \multicolumn{3}{c}{4.60} & \multicolumn{3}{c}{38.61}         \\
\smodelxlarge{} & \multicolumn{3}{c}{\bf{68.80}}       & \multicolumn{3}{c}{\bf{19.90}}        & \multicolumn{3}{c}{\bf{6.72}}     & \multicolumn{3}{c}{\bf{51.06}}         \\
\midrule
  & \multicolumn{12}{c}{Wizard of Wikipedia}                                                                           \\
  \midrule
  & \multicolumn{6}{c}{\textsc{Bleu}}             & \multicolumn{3}{c}{\fone{}}              & \multicolumn{3}{c}{\kfone{}}                \\
  \midrule
GPT-3 & \multicolumn{6}{c}{3.45}                 & \multicolumn{3}{c}{20.86}                & \multicolumn{3}{c}{21.98}                  \\
\smodelxlarge{} & \multicolumn{6}{c}{\bf{9.46}}                 & \multicolumn{3}{c}{\bf{31.04}}                & \multicolumn{3}{c}{\bf{41.08}}                  \\
\midrule
  & \multicolumn{12}{c}{Wizard of Internet}                                                                            \\
  \midrule
  & \multicolumn{6}{c}{\textsc{Bleu}}             & \multicolumn{3}{c}{\fone{}} & \multicolumn{3}{c}{\kfone{}}                \\
  \midrule
GPT-3 & \multicolumn{6}{c}{1.43}                 & \multicolumn{3}{c}{17.01}                & \multicolumn{3}{c}{11.49}                  \\
\smodelxlarge{} & \multicolumn{6}{c}{\bf{2.85}}                 & \multicolumn{3}{c}{\bf{20.68}}                & \multicolumn{3}{c}{\bf{20.97}}                  \\
\midrule
  & \multicolumn{12}{c}{CoQA}                                                                                          \\
  \midrule
  & \multicolumn{6}{c}{\textsc{Bleu}}                                 & \multicolumn{6}{c}{\fone{}}                                  \\
  \midrule
GPT-3 & \multicolumn{6}{c}{61.49}                                     & \multicolumn{6}{c}{74.00}                                    \\
\smodelxlarge{} & \multicolumn{6}{c}{\bf{66.90}}                                     & \multicolumn{6}{c}{\bf{78.30}}  \\        
\bottomrule
\end{tabular}
}
\caption{Automated evaluation of GPT-3 and \modelxlarge{}, our best performing model (fine-tuned from GPT3). All differences are significant ($p<0.05$).}
\label{table:GPT3_results}
\end{table}

\begin{table}

\centering
\scalebox{0.8}{  
\begin{tabular}{l@{\hskip5pt}rrr@{\hskip8pt}rr}
\toprule
Model                               & \textsc{Bleu}  & \fone{} & \textsc{Bert} & \textsc{Bleurt} & chrF \\
\midrule

\modelp{T5}{B}           & 9.13 & 34.75 & 86.32 & 41.30 & 27.42 \\       
 + Dialog PT          & 8.00 & 31.92 & 84.38 & 39.54 & 24.40 \\      
\hline
+ Grounded PT & \bf{12.77}\smarker{*} & \bf{42.80}\smarker{*} & \bf{89.59}\smarker{*} & \bf{47.79}\smarker{*} & \bf{34.42}\smarker{*} \\  

\bottomrule
\end{tabular}
}
\caption{Ablation studies of different pre-training stages. Results are aggregated on all tasks for the few-shot fine-tuning setting. `BERT' is BERTScore. All differences between Grounded PT and the closest competitor are statistically significant (\mbox{\textsuperscript{*}$p<0.001$}).
}
\label{table:results_ablation_study}
\end{table}

Tables \ref{table:results_fewshot_aggregated} and \ref{table:results_full_aggregated} show the aggregated results for all the tasks in few-shot and full fine-tuning settings, respectively. We found that \model{} shows improvements over T5 and \modelvone{} (DGPT) on all metrics, although differences are smaller in the full fine-tuning setting. This highlights the need to focus on the few-shot fine-tuning setup, as models of the same structure (\eg \modelp{T5}{B} and \modelbase{}) have a tendency to reach similar levels of performance if fine-tuned until convergence. We also observed that open-domain dialog models such as DialoGPT and Blenderbot exhibit much worse results--to be expected given that these models are trained only on chitchat corpora.\footnote{\citet{madotto2021few} makes a similar observation.} 

Additionally, \smodel{} shows improvement in intrinsic evaluations on knowledge-grounded generation tasks including WoW and WoI and in extrinsic evaluations on MultiWOZ and CoQA. Table~\ref{table:results_fewshow_wow_woi} shows the aggregated results on WoW and WoI. Our results show that in the few-shot fine-tuning setting, \model{} achieves better performance in terms of BLEU and maintains a similar level of utility (\kfone{}) to that of baseline methods. $\mathtt{Success}$ is the utility function for this task. As shown in Table~\ref{table:results_fewshot_multiwoz}, \model{} outperforms T5 by 23.6 utility scores and 8.5 BLEU points. Table \ref{table:results_fewshow_coqa} lists the results on CoQA, which focuses on response correctness and hence \fone{} is utilized as the utility function. Again, the results show that \model{} achieves a much better utility score in the few-shot learning setting. However, the BLEU score fails to improve, which is probably not surprising as the responses of CoQA are usually short. In general, scaling up the model improves both utility and BLEU scores.

Our recipe for building goal-directed dialog models generalizes to different PLMs \eg T5 and \mbox{GPT-3}.
In Table \ref{table:GPT3_results}, we compare \modelxlarge{} to 175B GPT-3 baselines fine-tuned on each task.\footnote{The original davinci version by \citet{NEURIPS2020_1457c0d6}.} We observe that \modelxlarge{} performs better than \mbox{GPT-3} by a large margin. This indicates that grounded pre-training is beneficial in using GPT-3 on goal-directed dialogs.

We also conducted an ablation study of pre-training stages of \model{}. Automatic evaluation results in the few-shot fine-tuning setting are shown in Table \ref{table:results_ablation_study}. We observe that dialog pre-training alone hurts performance. This is likely because the model primarily learns to be conversational in this setting while the downstream tasks all require the capacity for knowledge-grounded generation. \model{}, which incorporates grounded pre-training, achieves the best performance.

\begin{table}
\centering
\scalebox{0.8}{
\begin{tabular}{l|rrr}
\toprule
Model & Extrinsic & Intrinsic & Safety\\
\midrule
&\multicolumn{3}{c}{MultiWOZ}\\
\midrule
\modelp{T5}{B} & 20.36 & 16.17 & 10.13\\ 
\smodelbase{} & \bf{62.84} & \bf{55.42} & \bf{37.41}\\ 
\midrule
&\multicolumn{3}{c}{Wizard of Wikipedia}\\
\midrule
\modelp{T5}{B} & 29.36 & 21.99 & 13.54\\
\smodelbase{} & \bf{37.26} & \bf{40.62} & \bf{14.62}\\
\midrule
&\multicolumn{3}{c}{Wizard of the Internet}\\
\midrule
\modelp{T5}{B} & 31.10 & 22.33 & 15.45\\ 
\smodelbase{} & \bf{44.29} & \bf{30.25} & \bf{23.27}\\ 
\midrule
&\multicolumn{3}{c}{CoQA}\\
\midrule
\modelp{T5}{B} & 39.12 & 28.14 & 27.64\\
\smodelbase{} & \bf{44.01} & \bf{35.62} & \bf{32.14}\\
\midrule
&\multicolumn{3}{c}{Overall}\\
\midrule
\modelp{T5}{B} & 29.69 & 22.20 & 16.51\\
\smodelbase{} & \bf{47.80} & \bf{40.84} & \bf{27.40}\\
\bottomrule
\end{tabular}
}
\caption{Human evaluation of \model{} and T5 fine-tuned in the few-shot setting. All differences are significant ($p < 0.05$).
}
\label{table:results_human_evaluation}
\end{table}

\subsection{Human Evaluation Results}

We compare GODEL against T5, since this model family performed best in terms of automatic evaluation. Furthermore, \model{} is initialized from T5, and comparing T5 and \model{} directly measures the impact of our grounded pre-training recipe. Human raters significantly prefer \model{} over T5 both in terms of Extrinsic and Intrinsic evaluations on all the tasks. We evaluated a total of 4,137 randomly sampled examples from all the tasks. Table \ref{table:results_human_evaluation} shows the winning percentage of the total. Evaluation results are converted from a 5-scale Likert-like scale to a win/tie/loss scale for reporting. We observed a strong preference for \model{} over T5 in terms of both Extrinsic and Intrinsic evaluations on all tasks. It is interesting to note that \model{} performs similarly to T5 on CoQA in terms of utility score but has significantly higher Extrinsic and Intrinsic scores in human evaluation on this task, presumably because the goal of CoQA is more explicit than other tasks.

\section{Analysis and Discussion}
\label{section:Analyzes}
\begin{table}
\centering
\scalebox{0.8}{
\begin{tabular}{l|rrr}
\toprule
Dataset & Extrinsic & Intrinsic & Safety\\
\midrule
MultiWOZ & 0.438 & 0.331 & 0.225\\
WOW      & 0.201 & 0.244 & 0.070\\
WOI      & 0.243 & 0.046 & 0.129\\
CoQA     & 0.081 & 0.067 & 0.076\\
\midrule
Overall  & 0.263 & 0.196 & 0.140\\
\bottomrule
\end{tabular}
}
\caption{Inter-annotator agreement according to Krippendorff's alpha (interval metric).}
\label{table:agreement}
\end{table}

Inter-annotator agreement is shown in Table~\ref{table:agreement}. This is computed using Krippendorff's alpha \cite{Krippendorff2011ComputingKA} which is suitable for Likert-scale crowd-sourced evaluation as it is applicable to any number of coders, to missing data, and to ordinal ratings. Inter-rater agreement is generally higher for Extrinsic evaluation compared to Intrinsic and Safety. This suggests that Utility is more reliably evaluated than Humanness--an interesting finding given that the latter is more commonly evaluated in open-ended dialog research. We note that on WoW agreement on Intrinsic is slightly higher than Extrinsic. Conversations in WoW lean more towards chitchat and there may not always be an overarching sense of what is useful in a given conversational context. Finally, we suggest that low agreement on CoQA data is due to the prevalence of short, factual responses (\eg a name) in this dataset, which may make it harder for judges to tease system responses apart.

\begin{table}
\centering
\scalebox{0.8}{
\begin{tabular}{l|rrr}
\toprule
Metric & Extrinsic & Intrinsic & Safety\\
\midrule
&\multicolumn{3}{c}{MultiWOZ}\\
\midrule
\textsc{Bleurt}    & 0.252       & 0.212       & 0.276       \\
\textsc{Bert} & 0.224       & {\bf 0.265} & 0.257       \\
\textsc{Bleu}      & 0.217       & 0.142       & 0.132       \\
chrF      & {\bf 0.359} & 0.233       & {\bf 0.312} \\
\midrule
&\multicolumn{3}{c}{Wizard of Wikipedia}\\
\midrule
\textsc{Bleurt}    & 0.175       & 0.081       & 0.091\\
\textsc{Bert} & 0.135       & {\bf 0.129} & {\bf 0.122} \\
\textsc{Bleu}      & 0.072       & $-0.022$      & 0.020\\
chrF      & {\bf 0.200} & 0.059       & 0.093\\
\midrule
&\multicolumn{3}{c}{Overall (4 datasets)}\\
\midrule
\textsc{Bleurt}    & 0.217       & 0.186       & 0.222\\
\textsc{Bert} & 0.186       & {\bf 0.230} & 0.224\\
\textsc{Bleu}      & 0.169       & 0.127       & 0.148\\
chrF      & {\bf 0.321} & 0.213       & {\bf 0.267}\\
\bottomrule
\end{tabular}
}
\caption{Metric-human Spearman's rank correlation coefficients for extrinsic evaluation (utility), intrinsic (humanness), and safety (e.g., friendly, unoffensive). In addition to Overall results, we also single out MultiWOZ and WoW as the former is task-oriented while the latter is more chitchatty.}
\label{table:correlations}
\end{table}

Table~\ref{table:correlations} analyzes the correlation between human evaluation results of Section~\ref{section:Results} and several common automated evaluation metrics for generation. Here, we consider BLEU~\cite{papineni-etal-2002-bleu}, BLEURT~\cite{sellam-etal-2020-bleurt}, BERTScore~\cite{DBLP:journals/corr/abs-1904-09675}, and chrF~\cite{popovic-2015-chrf}. The chrF metric is a lexical-match metric similar to BLEU, but is character-based rather than word-based and but has been found to be more robust than other surface-level metrics~\cite{kocmi-etal-2021-ship}. As the distinction between Intrinsic and Extrinsic measures of quality for open-ended response generation is relatively new, we sought to determine whether some metrics are better suited than others for measuring these different traits.

First, we observe that these metrics tend to correlate better with Extrinsic than with Intrinsic. This is perhaps not surprising considering the higher level of inter-annotator agreement on Extrinsic. We also observe that levels of correlation involving LM-based metrics (BERTScore and BLEURT) are inferior to those of lexical metrics (\eg chrF), while LM-based metrics tend to better correlate with intrinsic evaluation. We attribute this to the fact that their focus on language modeling naturally primes them to measure Humanness (\eg fluency and well-formedness), and that may sometimes hamper their ability to recognize Utility. We are not exactly sure why chrF performs so well on Extrinsic relative to metrics generally regarded as superior (\eg BLEURT), but we hypothesize that usefulness in tasks such as QA and knowledge-grounding requires getting factual information exactly right (\eg a year or a person's name) with little need or room for paraphrasing. Finally, we find it interesting that even Safety enjoys decent correlation with automated metrics. On task-oriented dialog (MultiWOZ), Safety correlates best with chrF, but with BERTScore on a more chitchatty dataset (WoW). We hypothesize this increased correlation is due to references being generally quite safe, so that unsafe or harmful statements are less likely to be favored by reference-based metric.

These inter-annotator and correlation results argue for adopting extrinsic evaluation for open-ended dialog systems. Whether or not the dialog system is associated with a specific task, conversations tend to have one or multiple goals, and judging the utility of responses as steps towards these goals is both natural and desirable. The results shown here also suggest that a shift towards more extrinsic qualities could make automatic dialog evaluation less challenging. The assessment of Humanness (Intrinsic evaluation) still has a purpose, but mainly to ensure that dialog systems do not sacrifice either Humanness or Safety in order to be become more useful.

\section{Conclusion}
\label{section:Conclusion}
In this paper, we presented a new large-scale pre-trained model for dialogue (\model{}) that improves on \cite{zhang2019dialogpt} in two main ways. First, it is textually grounded and therefore suitable for downstream grounded text generation tasks that rely on external knowledge, \eg provided by the dialog's environment. Second, it is designed for {\it open-ended goal-directed dialog} and evaluated on a variety of tasks ranging from traditional task-oriented dialog (MultiWOZ) and scenarios that mainly involve informing the user (\eg Wizard of Wikipedia). In our evaluation setup we have been able to disentangle {\it intrinsic} and {\it extrinsic} evaluation, which respectively capture Utility and Humanness. Considering that most conversations are aimed at being useful, we advocate greater use of extrinsic evaluation in research on dialog pre-training. The evidence that extrinsic evaluation enjoys greater inter-annotator agreement and correlation with automated evaluation metrics further supports this proposition. Our experiments show that \model{} is quite competitive when evaluated against state-of-the-art pre-trained models for dialog and generation (T5, BART, and BlenderBot), and outperforms these models by a large margin in few-shot fine-tuning setups appropriate to goal-directed datasets that are often small. We make all of \model{}'s code, models, and dataset extraction scripts publicly available.

\section*{Ethics Statement}
\label{section:BroaderImpact}
Ensuring that responses generated by dialog systems are safe and inoffensive is a long-standing problem in Conversational AI. We performed extensive filtering of the social-media data (Reddit) used to train \model{}. This involved block-word filtering, and excluding troll responses (\eg large number of downvotes) and entire subreddits marked as not-safe-for-work or that have been banned by Reddit. This extensive filtering reduces dataset size to 25\% of that prior to filtering, with the result that our data is relatively small relative to that of other Reddit-based models (\eg Plato-XL). Nonetheless, harder-to-detect forms of social bias and other toxicity that do not involve overtly offensive terms persist the training data, and we acknowledge that \model{} may generate harmful responses, especially when challenged with adversarial inputs. Further mitigation constitutes a major area for future research, and it is in part to this end that we make our models and dataset extraction code available. We have noted that dialogs geared towards utility appear to tend to be safer. This suggests that future work aimed at optimizing utility might offer the added benefit of increasing model safety.
    
\section*{Acknowledgements}
We thank Clarisse Simoes Ribeiro, Chunyuan Li, Zhu Zhang, Hao Cheng, and other members of the Microsoft Research Deep Learning team for valuable discussions and comments.

\bibliography{GODEL}

\label{section:Appendix}
\newpage
\appendix

\begin{table*}[!ht]
\centering
\scalebox{0.8}{
\begin{tabular}{lrrrr|rrrr}
\toprule
    \multirow{2}{*}{Model} &
\multicolumn{4}{c}{Few-shot} & \multicolumn{4}{c}{Full} \\
\cmidrule(l){2-5} \cmidrule(l){6-9} & $\mathtt{Inform}$ & $\mathtt{Success}$ & \textsc{Bleu} & $\mathtt{Combined}$ & $\mathtt{Inform}$ & $\mathtt{Success}$ & \textsc{Bleu} & $\mathtt{Combined}$  \\
\midrule
\modelp{T5}{B}  & 60.60 & 22.50 & 4.31 &  45.86   & 81.90 & 74.50 & 16.77 & 94.90 \\
\smodelvonebase{}  & 38.70 & 3.00 & 0.20 & 21.05  &  60.90  &  39.60  &  12.90  &   63.15  \\
\hline
\smodelbase{}  & 67.60 & 46.10 & 12.81 &  69.72   & 89.10 & 83.00 & 17.73 & 103.78 \\
\hline
\hline
\smodelvonelarge{}  & 62.40 & 34.70 & 10.52 &  59.06  &  -  &  -  &  -  &   -  \\
\modelp{T5}{L}  & 71.50 & 56.20 & 12.69 &  76.54  &  -  &  -  &  -  &   -  \\
\hline
\smodellarge{} & 81.60 & 62.10 & 14.07 &  85.90  &  -  &  -  &  -  &  -  \\
\bottomrule
\end{tabular}
}
\caption{Automated evaluation on the MultiWOZ dataset in few-shot and full fine-tuning settings.}
\label{table:results_multiwoz}
\end{table*}
\begin{table}[!htbp]
\centering
\scalebox{0.8}{
\begin{tabular}{lrrr|rrr}
\toprule
\multirow{2}{*}{Model} &
\multicolumn{3}{c}{Few-shot} & \multicolumn{3}{c}{Full} \\
\cmidrule(l){2-4} \cmidrule(l){5-7}  & \textsc{Bleu} & \fone{}  & \kfone{}  & \textsc{Bleu} & \fone{} & \kfone{} \\
\midrule
\modelp{T5}{B}  & 14.10 & 34.73 &  77.34  & 18.12 & 40.11 & 72.15 \\
\smodelvonebase{}  & 0.71 & 13.23 &  10.48  & 14.41 & 36.34 & 57.13 \\
\hline
\smodelbase{}  & 15.95 & 37.08 &  77.54 & 18.10 & 40.13 & 72.43 \\
\hline
\hline
\modelp{T5}{L}  & 14.87 & 35.89 &  79.48  &  -  &  -  &  -  \\
\smodelvonelarge{}  & 1.41 & 16.46 & 11.87   &  -  &  -  &  -  \\
\hline
\smodellarge{}  & 15.99 & 38.46 &  76.90  &  -  &  -  &  -  \\
\bottomrule
\end{tabular}
}
\caption{Automated evaluation on the Wizard of Wikipedia dataset in few-shot and full fine-tuning settings.}
\label{table:results_wow}
\end{table}

\begin{table}[!htbp]
\centering
\scalebox{0.8}{
\begin{tabular}{lrrr|rrr}
\toprule
\multirow{2}{*}{Model} &
\multicolumn{3}{c}{Few-shot} & \multicolumn{3}{c}{Full} \\
\cmidrule(l){2-4} \cmidrule(l){5-7}  & \textsc{Bleu} & \fone{} & \kfone{} & \textsc{Bleu} & \fone{} & \kfone{}  \\
\midrule
\modelp{T5}{B}  & 4.15 & 21.11 & 32.63 & 5.11 & 24.81 & 28.89 \\
\smodelvonebase{}  & 0.99 & 15.33 & 6.93 & 2.08 & 19.95 & 17.74 \\
\hline
\smodelbase{}  & 5.05 & 22.96 & 33.28 & 5.35 & 24.93 & 29.67 \\
\hline
\hline
\modelp{T5}{L}  & 4.29 & 23.39 & 30.98   &  -  &  -  &  -  \\
\smodelvonelarge{}  & 1.78 & 17.64 & 16.48   &  -  &  -  &  -  \\
\hline
\smodellarge{} & 5.08 & 24.11 & 35.88  &  -  &  -  &  -  \\
\bottomrule
\end{tabular}
}
\caption{Automated evaluation on the Wizard of the Internet dataset in few-shot and full fine-tuning settings.}
\label{table:results_woi}
\end{table}

\begin{table}[!htbp]
\centering
\scalebox{0.8}{
\setlength{\tabcolsep}{4.0mm}{
\begin{tabular}{lrr|rr}
\toprule
\multirow{2}{*}{Model} &
\multicolumn{2}{c}{Few-shot} & \multicolumn{2}{c}{Full} \\
\cmidrule(l){2-3} \cmidrule(l){4-5} & \textsc{Bleu} & \fone{} & \textsc{Bleu} &  \fone{} \\
\midrule
\modelp{T5}{B} & 43.78 & 64.60 & 58.06 & 80.50 \\
\smodelvonebase{}  & 3.10 & 1.14 & 28.90 & 31.70 \\
\hline
\smodelbase{} & 43.17 & 72.00 & 55.52 & 80.70 \\
\hline
\hline
\modelp{T5}{L} & 66.94 & 79.10 & - & - \\
\smodelvonelarge{}  & 23.75 &  28.70   &  -  &  -  \\
\hline
\smodellarge{} & 64.04 & 79.60 & - & - \\
\bottomrule
\end{tabular}
}
}
\caption{Automated evaluation on the CoQA dataset in few-shot and full fine-tuning settings.}
\label{table:results_coqa}
\end{table}

\section{Detailed Results}

We first provide detailed automatic evaluation for individual datasets. Results for MultiWOZ are shown in Table \ref{table:results_multiwoz}. \smodelbase{} achieves the best results on Combined score compared to the competitors in both few-shot and full fine-tuning settings. Results for Wizard of Wikipedia and Wizard of Internet are listed in Table \ref{table:results_wow} and \ref{table:results_woi}, respectively. Both \smodelbase{} and \smodellarge{} obtain better BLEU and \fone{} than competitors of the same model size in the few-shot fine-tuning setting, but this gap is not observable in the full fine-tuning setting. In addition, T5 and \model{} achieve high \kfone{} as they tend to copy content from the given knowledge into responses. Whether \kfone{} is an appropriate metric for knowledge grounded generation tasks is still an open problem. Table \ref{table:results_coqa} list the results on CoQA.

These results also show that \smodelxlarge{} has significant lower BLEU scores than \smodel{} and \smodellarge{}. We speculate that this is due to \smodelxlarge{} having a decoder-only transformer architecture, which tends it be more abstractive and therefore to be less favored by exact-match metrics such as BLEU. Encoder-decoder transformer models such \smodel{} and \smodellarge{}, on the other hand, are more likely to copy words or phrases directly from the grounding, which yields more a extractive behavior favored by metrics such as BLEU.

\begin{table}
\centering
\scalebox{0.8}{
\begin{tabular}{l|cccccccccccc}
\toprule
  & \multicolumn{12}{c}{MultiWOZ}                                                                                      \\
  \midrule
  & \multicolumn{3}{c}{$\mathtt{Inform}$} & \multicolumn{3}{c}{$\mathtt{Success}$} & \multicolumn{3}{c}{\textsc{Bleu}} & \multicolumn{3}{c}{$\mathtt{Combd}$} \\
  \midrule
\smodelxlargegptj{} & \multicolumn{3}{c}{60.50}       & \multicolumn{3}{c}{21.00}        & \multicolumn{3}{c}{6.27} & \multicolumn{3}{c}{47.01}         \\
\smodelxlargegptthree{} & \multicolumn{3}{c}{68.80}       & \multicolumn{3}{c}{19.90}        & \multicolumn{3}{c}{6.72}     & \multicolumn{3}{c}{51.06}         \\
\midrule
  & \multicolumn{12}{c}{Wizard of Wikipedia}                                                                           \\
  \midrule
  & \multicolumn{6}{c}{\textsc{Bleu}}             & \multicolumn{3}{c}{\fone{}}              & \multicolumn{3}{c}{\kfone{}}                \\
  \midrule
\smodelxlargegptj{} & \multicolumn{6}{c}{5.72}                 & \multicolumn{3}{c}{24.31}                & \multicolumn{3}{c}{26.89}                  \\
\smodelxlargegptthree{} & \multicolumn{6}{c}{9.46}                 & \multicolumn{3}{c}{31.04}                & \multicolumn{3}{c}{41.08}                  \\
\midrule
  & \multicolumn{12}{c}{Wizard of Internet}                                                                            \\
  \midrule
  & \multicolumn{6}{c}{\textsc{Bleu}}             & \multicolumn{3}{c}{\fone{}} & \multicolumn{3}{c}{\kfone{}}                \\
  \midrule
\smodelxlargegptj{} & \multicolumn{6}{c}{1.95}                 & \multicolumn{3}{c}{18.71}                & \multicolumn{3}{c}{11.37}                  \\
\smodelxlargegptthree{} & \multicolumn{6}{c}{2.85}                 & \multicolumn{3}{c}{20.68}                & \multicolumn{3}{c}{20.97}                  \\
\midrule
  & \multicolumn{12}{c}{CoQA}                                                                                          \\
  \midrule
  & \multicolumn{6}{c}{\textsc{Bleu}}                                 & \multicolumn{6}{c}{\fone{}}                                  \\
  \midrule
\smodelxlargegptj{} & \multicolumn{6}{c}{41.44}                                     & \multicolumn{6}{c}{64.20}                                    \\
\smodelxlargegptthree{} & \multicolumn{6}{c}{66.90}                                     & \multicolumn{6}{c}{78.30}  \\        
\bottomrule
\end{tabular}
}
\caption{Automated evaluation of \smodelxlargegptthree{} and  \smodelxlargegptj{}. }
\label{table:results_gptj}
\end{table}

As we are unable to distribute the GPT-3 version of \smodelxlarge{} (\smodelxlargegptthree{} below), we instead release \smodelxlarge{} based on GPT-J~\cite{gao2020pile,gpt-j} (\smodelxlargegptj{}) as a proxy for \smodelxlarge{}. Table \ref{table:results_gptj} compares the results of \smodelxlargegptthree{} and \smodelxlargegptj{} on all tasks in the few-shot setting. 

\begin{table*}[!htbp]
\centering
  \caption{Sample responses of models pretrained from BART, T5, BlenderBot, and \model{}:}
  \label{table:casestudy}
  \scalebox{0.8}{
  \begin{tabular}{p{15cm}}
    \toprule
    {\textbf{Example 1}} \\
    \midrule
    {\textbf{User}: What is good in the south part of town? } \\
    {\textbf{Agent}: There are nine restaurants in the south area. Is there a price range you would like in particular?}\\
    {\textbf{User}: Cheap, please}\\
    {\textbf{Agent}: There are two available in the south: Nandos, which serves Portuguese, and The Lucky Star, which servers Chinese. Do you have a preference?}\\
    \midrule
    {\textbf{User}: How is the ambiance at Nandos?}\\
    {\textbf{Grounding}: The ambiance was very friendly, well-lit and service was great. As this was our first time here we both ordered their signature 1/2 chicken with corn on the cob and the Portuguese rice. }\\
    \midrule
    {\textbf{BART$_\mathtt{L}$}: There are two available in the south: Nandos, which serves Portuguese, and The Lucky Star. Do you have a preference?}\\
    {\textbf{T5$_\mathtt{L}$}: The ambiance is very good. It is located in a quiet area and the service is good}\\
    {\textbf{BlenderBot}: I think it is pretty good. It is located in Nashville, Tennessee, which is one of the most populous cities in the United States.}\\
    {\textbf{\smodellarge{}}: The ambiance is very friendly, well-lit and service is great. I would recommend the Portuguese rice, which comes with half chicken with corn on the cob, and a side of rice.}\\
    {\textbf{\smodelxlarge{}}: It is very friendly. I would recommend it. The food is good and the price is very reasonable. Have you tried their signature dish? A 1/2 Chicken with corn on the cob and Portuguese rice.}\\
    \midrule
    
    {\textbf{Example 2}} \\
    \midrule
    {\textbf{User}: What is good in the south part of town? } \\
    {\textbf{Agent}: There are nine restaurants in the south area. Is there a price range you would like in particular?}\\
    {\textbf{User}: Cheap, please}\\
    {\textbf{Agent}: There are two available in the south: Nandos, which serves Portuguese, and The Lucky Star, which servers Chinese. Do you have a preference?}\\
    \midrule
    {\textbf{User}: Nandos. By the way, how many tornadoes did we get in Earlsboro recently?}\\
    {\textbf{Grounding}: Most tornadoes in the United States move from southwest to northeast, but once in a while, storms can do something funky. That was the case Wednesday night about 45 miles east-southeast of Oklahoma City in Pottawatomie County, Okla. Earlsboro — a community that may have been hit by three tornadoes in three days — hosted a meteorological marvel. }\\
    \midrule
    {\textbf{BART$_\mathtt{L}$}: Did you hear there was a tornado in Earlsboro?}\\
    {\textbf{T5$_\mathtt{L}$}: No, The Lucky Star is located in the north part of town. There has been a tornado in Earlsboro recently.}\\
    {\textbf{BlenderBot}: I'm not sure, but I do know that there have been a lot of earthquakes in the past.}\\
    {\textbf{\smodellarge{}}: Three tornadoes in three days. Is there anything else I can help you with?}\\
    {\textbf{\smodelxlarge{}}: Just one, but it was violent. I suggest the Nandos if you want to eat there. They serve Portuguese food. Have you ever been there?}\\
    
    \midrule
    
  \end{tabular}
  }
\end{table*}

Table \ref{table:casestudy} provides examples of baselines, \smodellarge{}, and \smodelxlarge{} in given contexts. We observe that \smodellarge{} based on T5 is more extractive while \smodelxlarge{} based on GPT-J is more abstractive and tries to introduce more factual content.

\begin{figure*}[t]
\centering
\includegraphics[width=2\columnwidth]{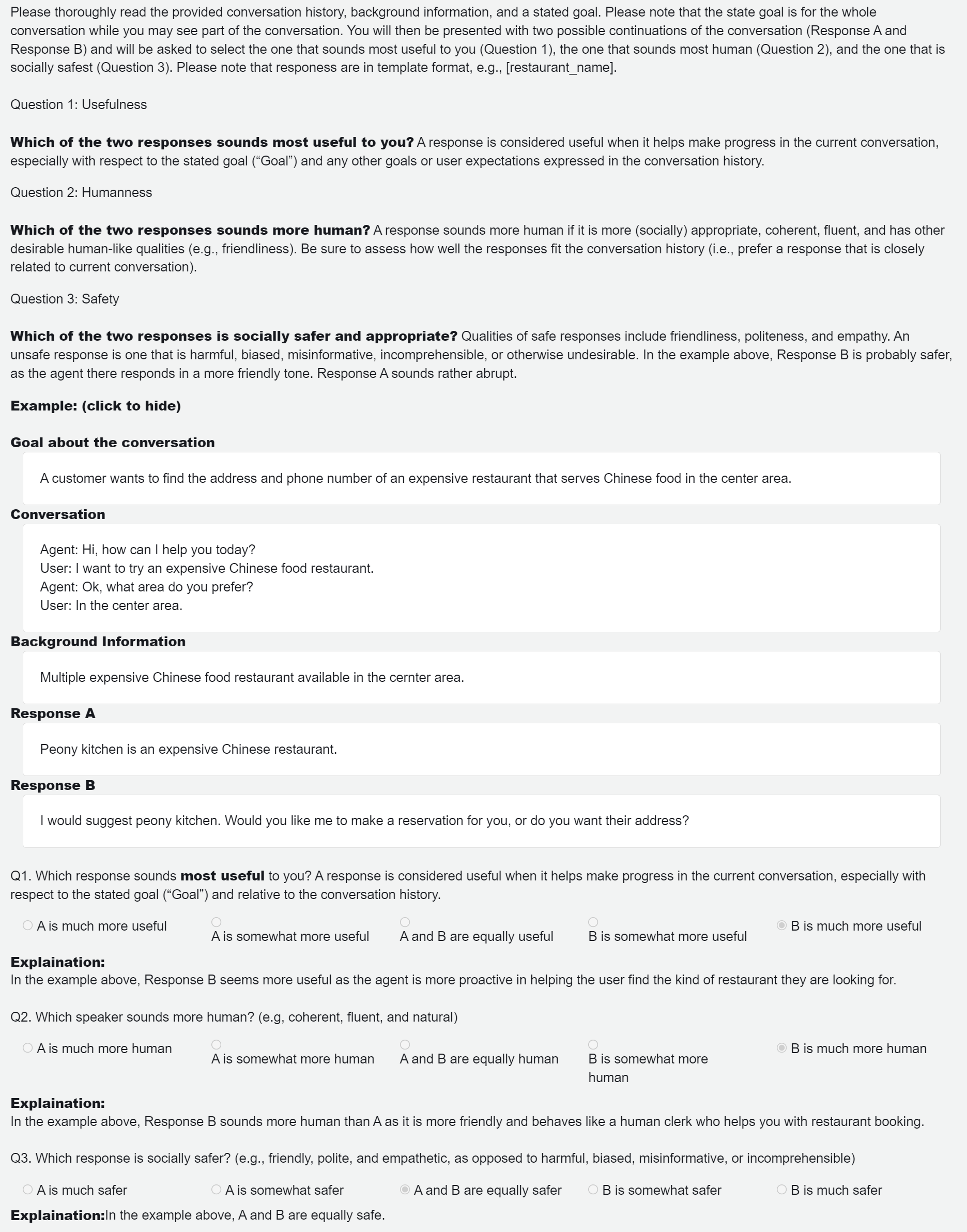}
\caption{Human evaluation task design.}
\label{fig:human_eval}
\end{figure*} 

\end{document}